\documentclass[10pt,twocolumn,letterpaper]{article}

\usepackage[pagenumbers]{wacv} % To force page numbers, e.g. for an arXiv version
\definecolor{wacvblue}{rgb}{0.21,0.49,0.74}
\usepackage[pagebackref,breaklinks,colorlinks,allcolors=wacvblue]{hyperref}

\usepackage{siunitx}

\def\wacvPaperID{1852} % *** Enter the WACV Paper ID here
\def\confName{WACV}
\def\confYear{2026}

\title{UniDiff: Parameter-Efficient Adaptation of Diffusion Models for Land Cover Classification with Multi-Modal Remotely Sensed Imagery and Sparse Annotations}

\author{
Yuzhen Hu\\
University of Houston\\
Houston, TX, USA\\
{\tt\small yhu34@uh.edu}
\and
Saurabh Prasad\\
University of Houston\\
Houston, TX, USA\\
{\tt\small saurabh.prasad@ieee.org}
}

\begin{document}
\maketitle
\begin{abstract}
Sparse annotations fundamentally constrain multimodal remote sensing: even recent state-of-the-art supervised methods such as MSFMamba are limited by the availability of labeled data, restricting their practical deployment despite architectural advances. ImageNet-pretrained models provide rich visual representations, but adapting them to heterogeneous modalities such as hyperspectral imaging (HSI) and synthetic aperture radar (SAR) without large labeled datasets remains challenging.
We propose UniDiff, a parameter-efficient framework that adapts a single ImageNet-pretrained diffusion model to multiple sensing modalities using only target-domain data. UniDiff combines FiLM-based timestep-modality conditioning, parameter-efficient adaptation of approximately 5\% of parameters, and pseudo-RGB anchoring to preserve pre-trained representations and prevent catastrophic forgetting. This design enables effective feature extraction from remote sensing data under sparse annotations.
Our results with two established multi-modal benchmarking datasets demonstrate that unsupervised adaptation of a pre-trained diffusion model effectively mitigates annotation constraints and achieves effective fusion of multi-modal remotely sensed data.
%On Berlin, UniDiff achieves 80.96\% overall accuracy, +4.04\% over MSFMamba, and on Augsburg, 93.08\% overall accuracy, +1.70\% over MSFMamba. These results show that unsupervised adaptation of a pretrained diffusion model effectively mitigates annotation constraints on the evaluated datasets and achieves effective multimodal fusion from remote sensing data.
\end{abstract}
    
\section{Introduction}
\label{sec:intro}

Dense pixel-level land-cover classification is a critical computer vision challenge with profound societal impact, enabling global climate monitoring, precision agriculture, disaster response, and urban planning at unprecedented scales. However, the high cost of pixel-level annotations severely limits the practicality of fully supervised multimodal fusion approaches at scale. Remote sensing captures complementary modalities such as hyperspectral imaging (HSI) and synthetic aperture radar (SAR), each encoding distinct physical properties. Effective fusion remains challenging due to differences across modalities~\cite{xu2018multisource}. Current approaches predominantly follow fully supervised paradigms; for example, recent methods like MSFMamba~\cite{zhang2025msfmamba} focus on architectural advances. While these sophisticated architectures achieve competitive performance with abundant labeled data, they remain constrained by annotation requirements—a limitation particularly pronounced in practical remote sensing applications, where substantial geospatial variability exacerbates sparse annotations~\cite{FPBC} and limits advanced supervised methods. In recent years, although large-scale labeled data have become available for multispectral and color satellite imagery \cite{van2021spacenet,xia2023openearthmap,garioud2023flair,clasen2024reben}, advanced sensing modalities such as hyperspectral and SAR imaging do not yet have large-scale labeled datasets. Additionally, owing to the significant variability of sensor and acquisition specifications of advanced active and passive sensors, creating homogeneous large-scale labeled datasets for such modalities is not possible at present.  

Additionally, foundation models such as SatMAE~\cite{cong2022satmae} and SpectralGPT~\cite{hong2024spectralgpt} that employ masked autoencoder~\cite{he2022masked} architectures for multispectral data have shown promise, while GeodiffNet-F~\cite{hu2025label} demonstrates the potential of diffusion models for remote sensing classification. However, many of these approaches are largely limited to single modalities and cannot fully address the challenges posed by heterogeneous sensing modalities.

Diffusion models~\cite{ho2020denoising,song2020score,sohl2015deep} offer a promising solution for multimodal fusion. Pretrained diffusion models achieve state-of-the-art performance in few-shot segmentation on natural images~\cite{baranchuk2021label}. Unlike other self-supervised methods such as DINO~\cite{caron2021emerging}, which struggle on textureless inputs, diffusion models leverage multi-scale noise training to capture spatial layouts effectively~\cite{zhang2023tale}. This allows them to extract meaningful features from challenging remote sensing data without relying on texture information~\cite{hu2025label}. Furthermore, their timestep conditioning mechanism~\cite{perez2018film} provides a natural pathway for multimodal extension through joint timestep–modality conditioning, enabling unified processing of heterogeneous sensing modalities without separate encoders.

Direct application of ImageNet-pretrained diffusion models to remote sensing, however, faces substantial domain gaps. Existing cross-domain adaptation methods often require extensive parameter updates, increasing computational costs and risking catastrophic forgetting~\cite{kirkpatrick2017overcoming}. Parameter-eficient approaches such as LoRA~\cite{hu2022lora} and DiffFit~\cite{xie2023difffit}, while more efficient, rely on static modifications that may lack expressiveness for heterogeneous multimodal processing.

To address these challenges, we extend FiLM-based conditioning from semantic to sensor diversity, introducing joint timestep–modality conditioning that allows a single backbone to process multiple modalities and produce modality-specific features (e.g., HSI-optimized and SAR-optimized) while keeping the pretrained backbone frozen. Additionally, we propose a pseudo-RGB anchoring strategy to maintain alignment with ImageNet statistics during adaptation, mitigating catastrophic forgetting.

In summary, our contributions are:

\begin{enumerate}
    \item \textbf{Unsupervised parameter-efficient multimodal adaptation.} We adapt a single ImageNet-pretrained diffusion model to heterogeneous modalities (HSI and SAR) using only unlabeled target data, updating just \textasciitilde5\% of parameters. The adapted model learns aligned multimodal representations within a shared backbone, enabling effective fusion under sparse supervision.

    \item \textbf{Joint timestep–modality conditioning mechanism.} We extend diffusion conditioning from timestep-only to joint timestep–modality conditioning via FiLM layers that learn $\gamma(t,m)$ and $\beta(t,m)$. This enables dynamic, modality-specific feature extraction within a shared backbone while preserving pretrained generalization.

    \item \textbf{Pseudo-RGB anchoring to prevent catastrophic forgetting.} Pseudo-RGB representations derived from HSI are used to preserve alignment with ImageNet statistics during adaptation. This anchoring mitigates catastrophic forgetting while enabling effective cross-modal feature learning.
\end{enumerate}

\section{Related Work}
\label{sec:LitRev}

\noindent\textbf{Multimodal Remote Sensing Fusion.} 
Combining hyperspectral imaging (HSI) and synthetic aperture radar (SAR) improves land-cover classification by exploiting complementary spectral and structural cues. Early approaches such as FusAtNet~\cite{mohla2020fusatnet}, S$^2$ENet~\cite{fang2022s2enet}, and DFINet~\cite{gao2021hyperspectral} used CNN-based late fusion, while recent designs—AsyFFNet~\cite{zhao2022asyffnet}, ExViT~\cite{yao2023exvit}, HCT~\cite{li2022hct}, MACN~\cite{wang2023macn}, and MSFMamba~\cite{zhang2025msfmamba}—incorporate attention mechanisms, state-space blocks, and memory enhancements to capture long-range dependencies. Despite architectural advances, these methods remain predominantly fully supervised, constraining scalability when dense annotations are expensive across large geographical regions. This motivates pretrained approaches that reduce annotation dependence and improve generalization.

\noindent\textbf{Pretrained Models in Remote Sensing.}
SatMAE~\cite{cong2022satmae} and SpectralGPT~\cite{hong2024spectralgpt} adopt self-supervised masked autoencoders for multispectral imagery, while GeoDiffNet-F~\cite{hu2025label} leverages ImageNet-pretrained diffusion models for HSI classification. Such pretrained features enable processing at larger spatial scales without proportional annotation costs, but these methods remain single-modality. In particular, GeoDiffNet-F relies on converting HSI into pseudo-RGB representations, making extension to disparate modalities like SAR challenging due to their different sensor characteristics.

\noindent\textbf{Diffusion Models for Feature Extraction.}  
Pretrained diffusion models provide robust spatial representations for dense prediction tasks, showing strong performance even with limited labels~\cite{baranchuk2021label}. They remain effective in challenging scenarios where other self-supervised methods may struggle~\cite{zhang2023tale}—a property especially relevant for remote sensing imagery with subtle or ambiguous textures, such as SAR or low-resolution multispectral data. Domain-specific efforts include DDPM-CD~\cite{bandara2022ddpm} for change detection and SpectralDiff~\cite{chen2023spectraldiff} for HSI benchmarks, although training diffusion models from scratch is computationally expensive and constrained by dataset scale. Despite these challenges, leveraging ImageNet-pretrained diffusion models for multimodal fusion remains largely unexplored.

\noindent\textbf{Parameter-Efficient Adaptation.}  
Methods such as LoRA~\cite{hu2022lora,gandikota2024concept} enable efficient adaptation through low-rank weight updates, while DiffFit~\cite{xie2023difffit} shows that updating a small subset of parameters—often normalization-related (e.g., scale and shift)—can yield effective domain adaptation. Recent work also demonstrates that diffusion model transferability varies across denoising timesteps, suggesting that adaptation strategies should account for these timestep-dependent differences~\cite{zhong2024diffusion}. These insights motivate our design of a joint timestep–modality conditioning scheme that adapts diffusion features efficiently without full fine-tuning.

\noindent\textbf{FiLM and Adaptive Conditioning.}  
Feature-wise Linear Modulation (FiLM)~\cite{perez2018film} introduced feature-wise affine conditioning, unifying concepts such as Conditional Batch Normalization and AdaIN~\cite{huang2017arbitrary}. Diffusion models leverage similar mechanisms: ADM employs Adaptive Layer Normalization for class conditioning~\cite{dhariwal2021diffusion}, while DiT extends adaptive normalization to transformer backbones~\cite{peebles2023scalable}. Building on this paradigm, we reinterpret class-conditioning pathways as modality-aware signals, enabling parameter-efficient adaptation for multimodal remote sensing fusion.

\section{Method}
\begin{figure*}[t]
    \centering
    \includegraphics[width=1\linewidth]{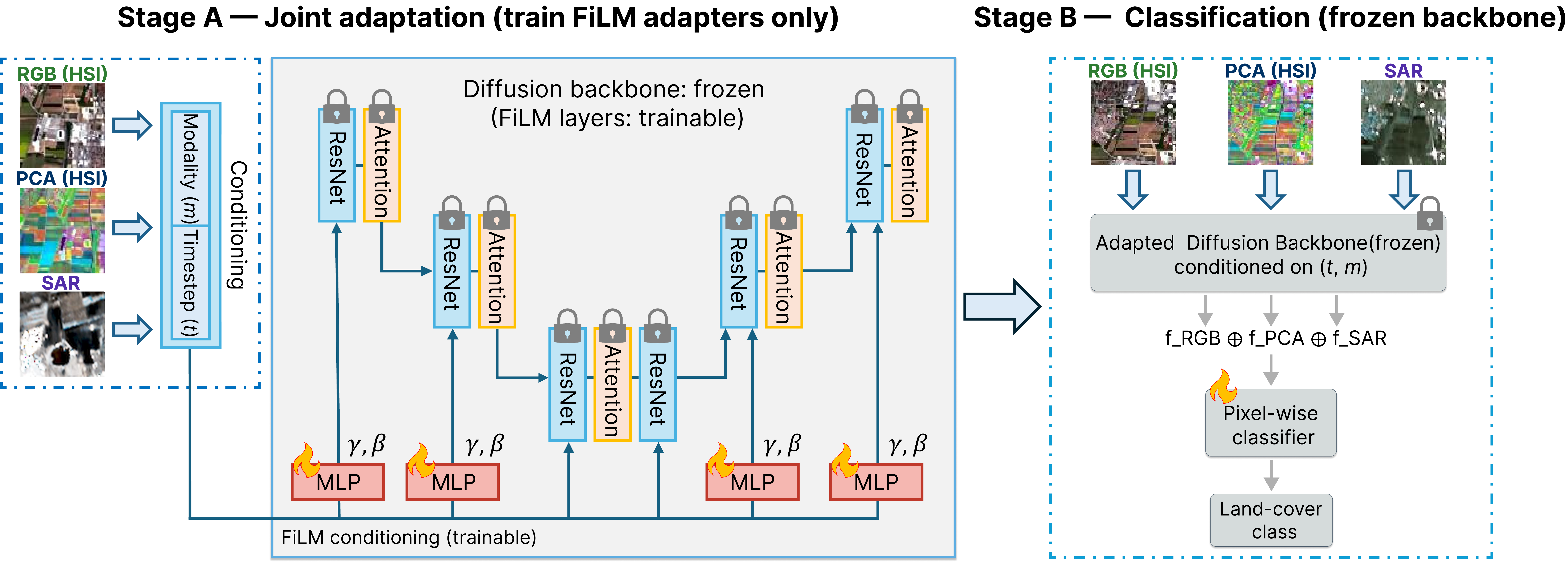} % Adjust the path and width as needed

\caption{
Two-stage parameter-efficient diffusion adaptation for multimodal remote sensing classification using ImageNet-pretrained models. Stage A: Joint adaptation using patch-based FiLM conditioning with pseudo-RGB anchoring. Stage B: Pixel-wise classification on co-registered multimodal features.
}

    \label{fig:unidiff_diagram}
\end{figure*}

\subsection{Problem Statement}
We study the problem of adapting pretrained diffusion models for multi-source remote sensing pixel-level classification under sparse supervision. 

Given multimodal inputs (HSI and SAR), we construct standardized 3-channel representations compatible with RGB-pretrained backbones: 
$\tilde{X}_{\text{pRGB}}$ (HSI-derived pseudo-RGB via band selection), 
$\tilde{X}_{\text{PCA}}$ (PCA-reduced HSI), and 
$\tilde{X}_{\text{SAR}}$ (Pauli-decomposed SAR), 
as illustrated in \cref{Berlin_UniDiff_MM_visual_compara} and \cref{UniDiff_Augsburg_Visual}.

This yields standardized inputs $\mathbf{X}^m \in \mathbb{R}^{H \times W \times 3}$, where 
$m \in \{\text{pRGB}, \text{PCA}, \text{SAR}\}$ indexes the representation type.  
Sparse supervision is available in the form of pixel-level labels:
\begin{equation}
\mathcal{Y} = \{(i,j,y_{ij}) : (i,j) \in \Omega\},
\label{eq:sparse_labels}
\end{equation}
where $|\Omega|$ accounts for less than 1\% of pixels (0.97\% for Augsburg and 0.61\% for Berlin; see \cref{tab:Augsburg_label_distribution} and \cref{tab:label_distribution_berlin}). 
\textbf{The goal is to classify each pixel into semantic land cover classes (e.g., forest, residential, industrial areas).}

Our objective is to leverage pretrained diffusion models for effective multimodal remote sensing classification under sparse supervision. To bridge the domain gap, we adapt them through parameter-efficient dynamic conditioning to extract representation-specific discriminative features while retaining the spatial priors learned from large-scale RGB pretraining.

\subsection{Method Overview}
Our approach consists of two stages: joint adaptation across representations followed by feature extraction and classification, as illustrated in \cref{fig:unidiff_diagram}.

\textbf{Stage A -- Joint Adaptation for Dense Feature Learning.}
We adapt a pretrained diffusion model to remote sensing data by repurposing its conditioning mechanism for representation-aware adaptation (see \cref{sec:modality_aware_conditioning}). 
Different noise levels require different adaptations: early timesteps (high noise) benefit from different transformations than late timesteps (low noise). 
We therefore generate conditioning parameters $\gamma(t,m)$ and $\beta(t,m)$ that jointly depend on timestep $t$ and input representation $m \in \{\text{pRGB}, \text{PCA}, \text{SAR}\}$, allowing dynamic adaptation throughout the denoising process. 
All three representations are trained jointly, with HSI-derived pRGB used as an anchor to stabilize training (see \cref{sec:anchor_adaptation}). 
Importantly, this stage is parameter-efficient: only the lightweight conditioning networks are updated, while the large diffusion backbone remains frozen, as illustrated in \cref{fig:unidiff_diagram}.

\textbf{Stage B --  Feature Extraction and Classification.}
The adapted model, kept fully frozen, serves as a universal feature extractor. Pixel-wise classification is performed with fewer than 1\% labeled data, without task-specific backbone training. 

\textbf{Key Distinction.}
Unlike static parameter-efficient methods (e.g., LoRA~\cite{hu2022lora}, DiffFit~\cite{xie2023difffit}) that apply fixed transformations, our timestep--representation conditioning adapts dynamically across both the denoising trajectory and input type, which is crucial for heterogeneous remote sensing data.
\subsection{Modality-Aware Conditioning via Repurposed AdaGN}
\label{sec:modality_aware_conditioning}
Modern diffusion models inject conditioning signals through Adaptive Group Normalization (AdaGN) layers ~\cite{dhariwal2021diffusion} that apply scale--shift transformations:
\begin{equation}
\text{AdaGN}(h, c) = \gamma_c \cdot \text{GroupNorm}(h) + \beta_c.
\end{equation}
We extend this FiLM-style conditioning mechanism ~\cite{perez2018film} to incorporate modality information by introducing learned embeddings $e_m$ for each representation type $m \in \{\text{pRGB}, \text{PCA}, \text{SAR}\}$:
\begin{align}
\gamma(t,m) &= \text{MLP}_{\gamma}(e_t + e_m), \\
\beta(t,m)  &= \text{MLP}_{\beta}(e_t + e_m),
\end{align}
where $e_t$ is the timestep embedding. This joint conditioning enables dynamic, context-sensitive adaptation -- i.e., each representation receives distinct normalization parameters that vary across denoising timesteps.

\textbf{Novel Contribution.} 
Existing diffusion approaches for remote sensing are typically trained from scratch for single modalities, while parameter-efficient adaptation methods (LoRA, DiffFit) apply static transformations that cannot account for heterogeneous sensor statistics. We introduce \emph{joint timestep--modality conditioning}, extending diffusion conditioning from timestep-only $\gamma(t)$ to $\gamma(t,m)$ and $\beta(t,m)$ via FiLM layers. This enables dynamic adaptation that varies across both denoising timesteps and input modalities, allowing a single ImageNet-pretrained backbone to extract modality-specific features. Unlike static adaptation methods, which struggle with statistically divergent modalities (e.g., natural RGB-like HSI vs.\ synthetic aperture radar), our approach updates only $\sim$5\% of parameters through lightweight conditioning networks while keeping the pretrained backbone frozen, thus preserving learned spatial priors during multimodal adaptation.

\textbf{Representation-Specific Feature Learning.} 
The joint conditioning mechanism enables modality-specific normalization parameters within the shared backbone, supporting distinct transformations for each representation: pseudo-RGB preserves natural image statistics, PCA-reduced HSI captures spectral variations, and SAR accounts for synthetic aperture structures. This modality-aware adaptation occurs while maintaining unified spatial processing through the frozen backbone, enabling effective cross-modal feature learning without separate encoders for each sensor type.

\subsection{Anchor-Based Adaptation from Pretraining}
\label{sec:anchor_adaptation}
\textbf{Anchor-Based Drift Prevention.}  
Diffusion models that process multiple modalities through a shared backbone are vulnerable to representation drift, where adaptation to divergent inputs can degrade performance below pretrained baselines. Although continual learning approaches mitigate forgetting by replaying prior task data~\cite{kirkpatrick2017overcoming}, this is infeasible in our setting. Instead, we employ HSI-derived pRGB as an anchor, constraining drift while enabling domain-specific adaptation — a practical strategy for heterogeneous remote sensing modalities with divergent statistics.

\textbf{Design Rationale.}  
Pseudo-RGB serves a dual role: a processing modality (alongside PCA-reduced HSI and SAR) and a distributional anchor to ImageNet statistics. By leveraging the statistical similarity between HSI-derived pseudo-RGB and natural RGB images that ImageNet diffusion models were pretrained on, pseudo-RGB helps preserve alignment with pretrained spatial priors during adaptation. This stabilizing effect is particularly important when training jointly with statistically divergent modalities such as SAR, which lack RGB-like properties and otherwise risk inducing representation drift in the shared backbone.

The model is jointly trained using the diffusion denoising objective:
\begin{equation}
\mathcal{L}_{\text{denoise}} =
\mathbb{E}_{x_0, t, m, \epsilon}\left[
\left\| \epsilon - \epsilon_\theta(x_t, t, m) \right\|_2^2
\right],
\end{equation}
where $m \in \{\text{pRGB}, \text{PCA}, \text{SAR}\}$ represents the three representations processed jointly, with pRGB providing distributional stability throughout adaptation.

\subsection{Downstream Classification}

\textbf{Stage B: Frozen Feature Extraction.}  
Following Stage A adaptation, the diffusion backbone remains frozen and serves as a feature extractor (\cref{fig:unidiff_diagram}, Stage B) that produces pixel-wise modality-specific features conditioned on both denoising timestep $t$ and modality type $m \in \{\text{pRGB}, \text{PCA}, \text{SAR}\}$. During training, these conditioned features are used with sparse pixel-level annotations to supervise a lightweight classifier. At inference, the same conditioning mechanism enables dense prediction across all pixels for each of the three modality representations. This design ensures that performance reflects the representational quality gained through joint timestep–modality adaptation, rather than additional model capacity.

\textbf{Classifier Architecture.}  
A lightweight MLP head maps frozen diffusion features to pixel-wise class predictions. We freeze the backbone to avoid overfitting under sparse supervision, as fine-tuning large pretrained models in low-label regimes can degrade representations~\cite{yosinski2014transferable,kornblith2019better}---a concern particularly pronounced in remote sensing where labeled data is scarce~\cite{tuia2021recent}. This approach isolates the representational gains from Stage A adaptation rather than additional model capacity, validating the effectiveness of our joint timestep--modality conditioning framework.

\section{Experiments and Results}
\label{sec-exp}

\subsection{Datasets}
\label{subsec:datasets}

We evaluate our approach on two established multimodal benchmarks that provide complementary evaluation scenarios across different sensor resolution regimes: \textbf{Augsburg} and \textbf{Berlin}. These datasets test our method's robustness to different multimodal alignment strategies and spatial resolution characteristics. 

\textbf{Augsburg} (HSI: 180 bands, $332 \times 485$ pixels; SAR: VV/VH polarizations) represents a scenario where SAR data is downsampled to match coarser HSI resolution. Both modalities are co-registered to uniform $30\,\text{m}$ GSD, with the original higher-resolution SAR aligned to the native HSI spatial grid. This configuration tests our joint conditioning approach when SAR spatial detail is reduced to enable fusion.

\textbf{Berlin} (HSI: 244 bands, $797 \times 220$ pixels; SAR: Sentinel-1 VV/VH) represents the opposite scenario where HSI is upsampled to match finer SAR resolution at $13.89\,\text{m}$ GSD. The HSI data is resampled from its original $30\,\text{m}$ resolution to the higher-resolution SAR spatial grid (final dimensions: $1723 \times 476 \times 244$), testing our approach when HSI spatial detail is interpolated for fusion.

Following benchmark protocols from Hong et al.~\cite{hong2021multimodal}, we use the official train–test splits to ensure comparability with recent multimodal fusion studies~\cite{zhang2025msfmamba}. Both datasets exhibit severe class imbalance and label scarcity while providing dense pixel-level annotations (77,533 and 461,851 test pixels, respectively) that support comprehensive evaluation. This combination of dense annotations, class imbalance, and heterogeneous alignment provides a rigorous evaluation setting for sparse-label multimodal methods. Detailed class distributions are reported in \cref{tab:Augsburg_label_distribution} and \cref{tab:label_distribution_berlin}.

\begin{table}[h]
  \caption{Augsburg dataset: training and testing pixel counts}
  \label{tab:Augsburg_label_distribution}
  \centering
  \small
  \setlength{\tabcolsep}{3pt}
  \renewcommand{\arraystretch}{0.7}
  \begin{tabular}{lcr}
    \toprule
    \textbf{Class} & \textbf{Train Count} & \textbf{Test Count} \\
    \midrule
    Forest & 146 & 13,361 \\
    Residential Area & 264 & 30,065 \\
    Industrial Area & 21 & 3,830 \\
    Low Plants & 248 & 26,609 \\
    Allotment & 52 & 523 \\
    Commercial Area & 7 & 1,638 \\
    Water & 23 & 1,507 \\
    \midrule
    \textbf{Total} & \textbf{761} & \textbf{77,533} \\
    \bottomrule
  \end{tabular}
 \vspace{-10pt} % Uncomment if you need to save vertical space
\end{table}

\begin{table}[h]
  \centering
  \small
  \setlength{\tabcolsep}{4pt}
  \renewcommand{\arraystretch}{0.7}
  \caption{Berlin dataset: training and testing pixel counts}
  \label{tab:label_distribution_berlin}
  \begin{tabular}{lcr}
    \toprule
    \textbf{Class} & \textbf{Train Count} & \textbf{Test Count} \\
    \midrule
    Forest            & 443  & 54,511 \\
    Residential Area  & 423  & 268,219 \\
    Industrial Area   & 499  & 19,067 \\
    Low Plants        & 376  & 58,906 \\
    Soil              & 331  & 17,095 \\
    Allotment         & 280  & 13,025 \\
    Commercial Area   & 298  & 24,526 \\
    Water             & 170  & 6,502 \\
    \midrule
    \textbf{Total}    & \textbf{2,820} & \textbf{461,851} \\
    \bottomrule
  \end{tabular}
  % \vspace{-10pt} % Uncomment if you need to save vertical space
\end{table}

\vspace{-8pt}
\subsection{Implementation Details}

\textbf{Experimental Setup.} 
All experiments are conducted on NVIDIA GeForce RTX 4080 GPUs using PyTorch 2.2 with CUDA 12.1. Random seeds are fixed for reproducibility.

\textbf{Diffusion Backbone.} 
We use the ADM (U-Net) model pretrained on $64 \times 64$ ImageNet images~\cite{dhariwal2021diffusion,nichol2021improved}. 
The backbone remains frozen during adaptation, with only lightweight FiLM conditioning layers updated.

\textbf{Training Protocol.} 
We adopt a two-stage procedure: (\textit{i}) domain adaptation using the Adam optimizer (lr=0.003) by updating only FiLM layers (batch size 32, 2000 steps) while keeping the diffusion backbone frozen, and (\textit{ii}) MLP classifier training using Adam (lr=0.001, weight decay $5 \times 10^{-4}$) with batch size 64, 10 epochs, and early stopping. 
The classifier depth varies per dataset to match label set size.

\textbf{Data Processing.}  
Images are normalized using a 2nd--98th percentile stretch for all modalities: pseudo-RGB bands selected from HSI, the top three PCA components of HSI, and Pauli-decomposed SAR channels. 
The normalized images are then divided into $64 \times 64$ patches with $50\%$ overlap (stride $=32$). 
During inference, predictions from overlapping patches are merged by probability averaging to mitigate boundary artifacts.

\textbf{Layer and Timestep Selection.}  
We adopt the layer selection from GeoDiffNet-F~\cite{hu2025label} (layer 10 for Augsburg, layer 11 for Berlin) and determine timesteps through systematic ablation ($t=0$ for Augsburg, $t=300$ for Berlin). 
The corresponding ablation results that guided this choice are reported in Appendix~B.

\subsection{Main Results}
We evaluate UniDiff on the Berlin and Augsburg multimodal datasets, benchmarking against recent baselines. Following prior work, we report per-class recall, overall accuracy (OA), average accuracy (AA), and Cohen’s kappa as the primary metrics; additional F1 and IoU results are provided in Appendix~C.

\textbf{Multimodal Performance.} 
\cref{tab:Berlin_UniDiff_compare_sota,tab:Augsburg_UniDiff_compare_sota} compare UniDiff with baselines in the HSI+SAR setting.

\textit{Berlin Dataset.} UniDiff achieves 80.96\% OA, 68.08\% AA, and 70.05\% Kappa, surpassing MSFMamba (76.92\% OA, 64.88\% AA, 64.88\% Kappa) by +4.04\% OA, +3.20\% AA, and +5.17\% Kappa, showing the effectiveness of our parameter-efficient adaptation strategy.

\begin{table*}[t]
 \caption{Results on the \textbf{Berlin dataset} (per-class Recall). 
All methods are evaluated in the HSI+SAR multimodal setting. 
\textbf{UniDiff} shows consistent improvements over recent multimodal baselines. 
OA: Overall Accuracy, AA: Average Accuracy, Kappa: Kappa Coefficient (all reported in \%).}
\vspace{-6pt}
  \centering
  \scriptsize
  \setlength{\tabcolsep}{2pt}
  \renewcommand{\arraystretch}{0.85}
  \resizebox{\textwidth}{!}{%
    \begin{tabular}{lccccccccc}
      \toprule
      \textbf{Method} 
      & FusAtNet~\cite{mohla2020fusatnet} 
      & S$^2$ENet~\cite{fang2022s2enet} 
      & DFINet~\cite{gao2021hyperspectral} 
      & AsyFFNet~\cite{zhao2022asyffnet} 
      & ExViT~\cite{yao2023exvit} 
      & HCT~\cite{li2022hct} 
      & MACN~\cite{wang2023macn} 
      & MSFMamba~\cite{zhang2025msfmamba} 
      & \textbf{UniDiff (Ours)} \\
      \midrule
      Forest       & 86.24 & 83.27 & 82.04 & 88.35 & 84.71 & 83.18 & 82.21 & 82.39 & 75.76 \\
      Residential  & 91.38 & 72.07 & 77.78 & 74.44 & 76.19 & 78.80 & 78.16 & 82.05 & 88.66 \\
      Industrial   & 19.76 & 46.66 & 47.96 & 48.21 & 44.15 & 43.59 & 47.79 & 48.91 & 57.53 \\
      Low Plants   & 20.00 & 72.08 & 77.78 & 71.41 & 77.96 & 76.48 & 78.04 & 79.72 & 85.55 \\
      Soil         & 48.72 & 77.94 & 87.85 & 80.75 & 67.13 & 82.92 & 77.26 & 85.22 & 89.00 \\
      Allotment    & 38.89 & 70.62 & 44.75 & 48.35 & 61.95 & 56.46 & 66.07 & 71.25 & 37.21 \\
      Commercial   & 18.47 & 36.48 & 29.85 & 27.97 & 29.28 & 18.96 & 28.90 & 31.08 & 34.32 \\
      Water        & 29.61 & 54.64 & 60.09 & 75.16 & 56.98 & 44.16 & 50.22 & 38.48 & 76.58 \\
  
    \midrule
    OA (\%)    & 70.91 & 70.38 & 73.69 & 71.65 & 72.59 & 73.42 & 73.98 & 76.92 & \textbf{80.96} \\
    AA (\%)    & 44.13 & 64.22 & 63.51 & 64.33 & 62.39 & 60.57 & 63.58 & 64.88 & \textbf{68.08} \\
    Kappa (\%) & 51.07 & 57.73 & 61.02 & 58.74 & 59.72 & 60.29 & 61.19 & 64.88 & \textbf{70.05} \\

      \bottomrule
    \end{tabular}
  }
  \label{tab:Berlin_UniDiff_compare_sota}
\end{table*}

\textit{Augsburg Dataset.} UniDiff reaches 93.08\% OA, 71.68\% AA, and 89.97\% Kappa, outperforming MSFMamba (91.38\% OA, 63.31\% AA, 87.45\% Kappa) by +1.70\% OA, +8.37\% AA, and +2.52\% Kappa, confirming robustness across regions.

\begin{table*}[t]

\caption{Comparison on the \textbf{Augsburg dataset} (per-class Recall). 
\textbf{UniDiff (HSI+SAR)} achieves consistent improvements over recent multimodal baselines. 
OA: Overall Accuracy, AA: Average Accuracy, Kappa: Kappa Coefficient (all reported in \%).}
\vspace{-6pt}
  \centering
  \scriptsize
  \setlength{\tabcolsep}{2pt}
  \renewcommand{\arraystretch}{0.85}
  \resizebox{\textwidth}{!}{%
    \begin{tabular}{lccccccccc}
      \toprule
      \textbf{Method} & FusAtNet~\cite{mohla2020fusatnet} & S$^2$ENet~\cite{fang2022s2enet} & DFINet~\cite{gao2021hyperspectral} & AsyFFNet~\cite{zhao2022asyffnet} & ExViT~\cite{yao2023exvit} & HCT~\cite{li2022hct} & MACN~\cite{wang2023macn} & MSFMamba~\cite{zhang2025msfmamba} & \textbf{UniDiff (Ours)} \\ 
      \midrule
      Forest       & 94.34 & 98.10 & 97.38 & 97.56 & 90.04 & 97.26 & 97.84 & 97.17 & 97.28 \\
      Residential  & 92.56 & 99.08 & 98.37 & 99.16 & 95.44 & 98.63 & 98.73 & 98.15 & 97.87 \\
      Industrial   & 47.70 & 12.19 & 61.31 & 61.38 & 34.58 & 36.61 & 43.94 & 50.26 & 58.02 \\
      Low Plants   & 85.96 & 91.78 & 92.63 & 83.47 & 90.68 & 93.54 & 94.11 & 95.52 & 98.01 \\
      Allotment    & 49.33 & 45.12 & 49.33 & 42.64 & 51.82 & 51.43 & 52.39 & 53.35 & 85.47 \\
      Commercial   & 11.52 & 1.22  & 3.54  & 5.58  & 28.63 & 7.97  & 6.80  & 2.63  & 13.31 \\
      Water        & 45.27 & 24.09 & 26.61 & 45.81 & 17.65 & 46.34 & 48.54 & 49.97 & 51.82 \\
      \midrule
      OA (\%)      & 85.31 & 88.12 & 90.66 & 88.25 & 86.65 & 90.33 & 91.06 & 91.38 & \textbf{93.08} \\
      AA (\%)      & 62.93 & 53.08 & 61.31 & 62.23 & 58.41 & 61.68 & 63.19 & 63.31 & \textbf{71.68} \\
      Kappa (\%)   & 86.33 & 86.47 & 86.47 & 83.13 & 80.79 & 86.03 & 87.06 & 87.45 & \textbf{89.97} \\
      \bottomrule
    \end{tabular}
  }
  \label{tab:Augsburg_UniDiff_compare_sota}
\end{table*}

\textbf{Scalability.} 
In contrast to baselines constrained to fixed multimodal inputs, UniDiff accommodates both HSI-only and HSI+SAR configurations within a unified framework. 
For example, UniDiff achieves 74.96\% OA on Berlin and 92.09\% OA on Augsburg with HSI-only inputs (Appendix~B), while further improving to 80.96\% and 93.17\% OA respectively when fusing HSI+SAR. 
This flexibility enables deployment under incomplete data conditions, avoiding the need to relearn modality relationships from scratch as in fully supervised methods.

\begin{figure}[!h]
    \centering
    \includegraphics[width=0.9\linewidth]{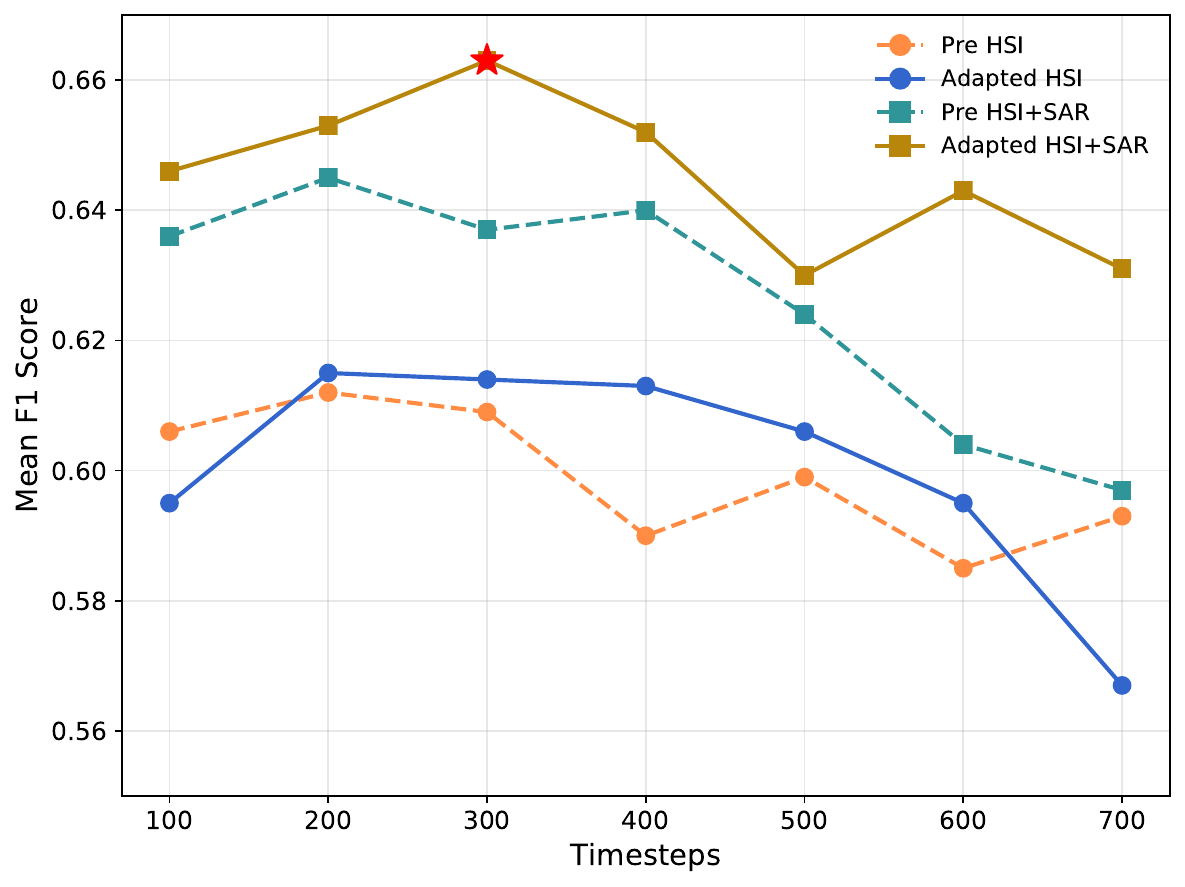} 
    \caption{Adaptation benefits across denoising timesteps on the Berlin dataset (Mean F1 score). 
The adapted HSI+SAR configuration consistently outperforms pretrained baselines, 
with peak performance observed at timestep 300.}

    \label{fig:berlin_timestep_f1}
\end{figure}

\textbf{Adaptation Stability Analysis.} 
We examine performance across denoising timesteps to assess the effect of our joint condition adaptation. 
\cref{fig:berlin_timestep_f1} shows mean F1 scores on the Berlin dataset, comparing adapted configurations with non-adapted counterparts that directly reuse ImageNet-pretrained features (Pre HSI and Pre HSI+SAR). 
Across timesteps 100--700, the adapted variants achieve consistently higher performance, with peak accuracy at timestep~300. 
These results suggest that adaptation provides clear and stable benefits over directly applying ImageNet-pretrained features, helping to mitigate the domain gap between natural images and multimodal remote sensing data. 
Comprehensive timestep analyses for both datasets are reported in Appendix~B.

\begin{figure*}[t]
    \centering
    % Compact, height-fixed, evenly spaced subfigures
    \begin{subfigure}[b]{0.15\textwidth}
        \includegraphics[height=0.46\textheight]{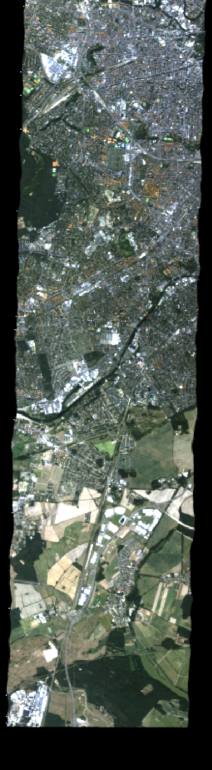}
        \caption{RGB}
    \end{subfigure}
    \hspace{0.3em}
    \begin{subfigure}[b]{0.15\textwidth}
        \includegraphics[height=0.46\textheight]{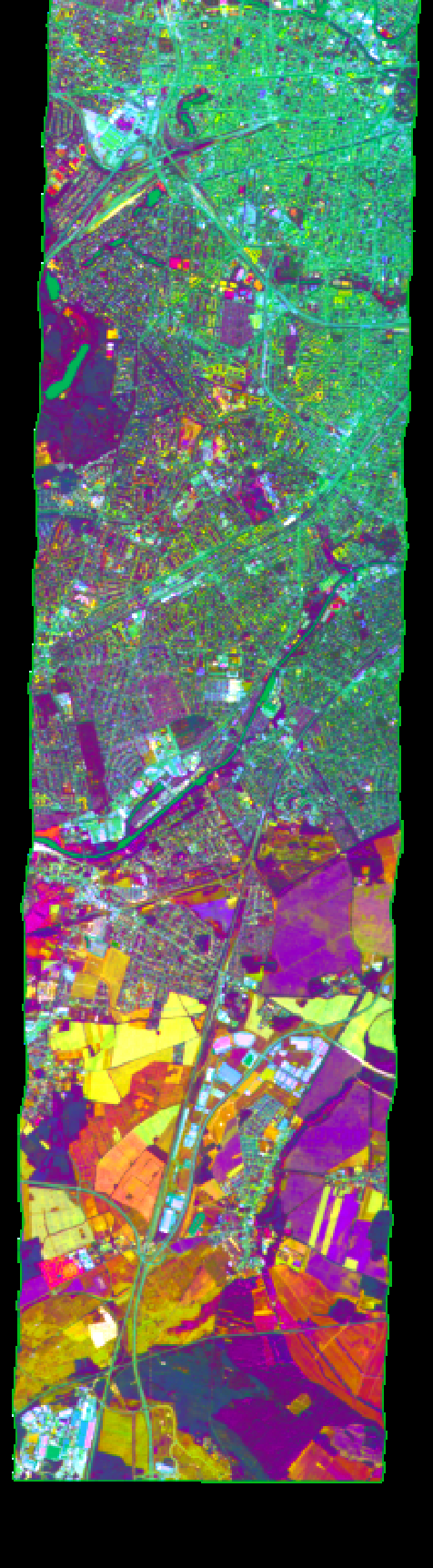}
        \caption{PCA}
    \end{subfigure}
    \hspace{0.3em}
    \begin{subfigure}[b]{0.15\textwidth}
        \includegraphics[height=0.46\textheight]{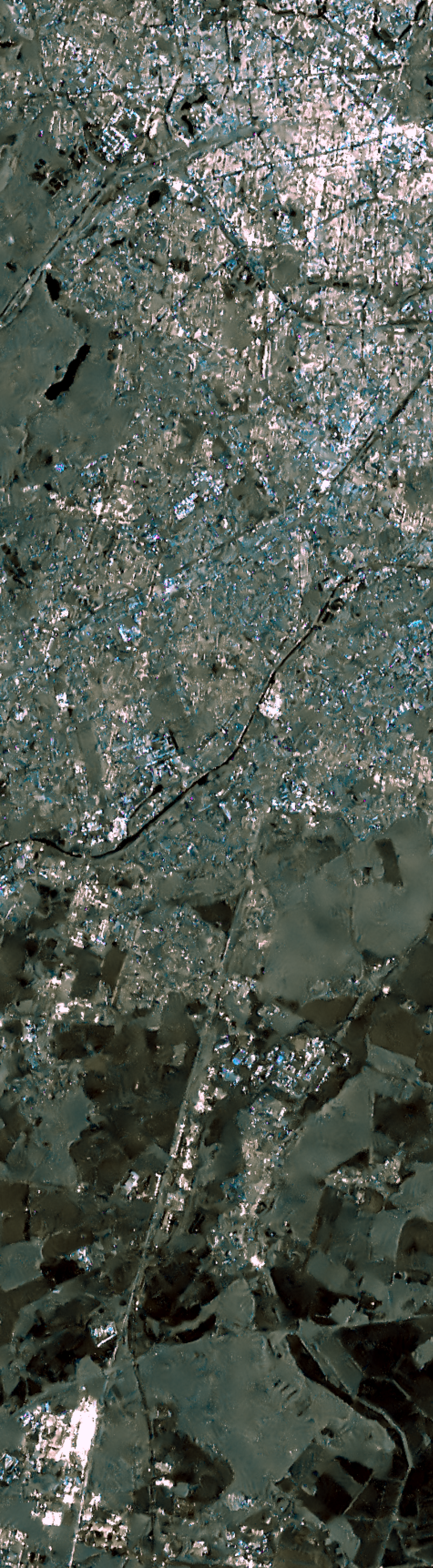}
        \caption{SAR}
    \end{subfigure}
    \hspace{0.3em}
    \begin{subfigure}[b]{0.15\textwidth}
        \includegraphics[height=0.46\textheight]{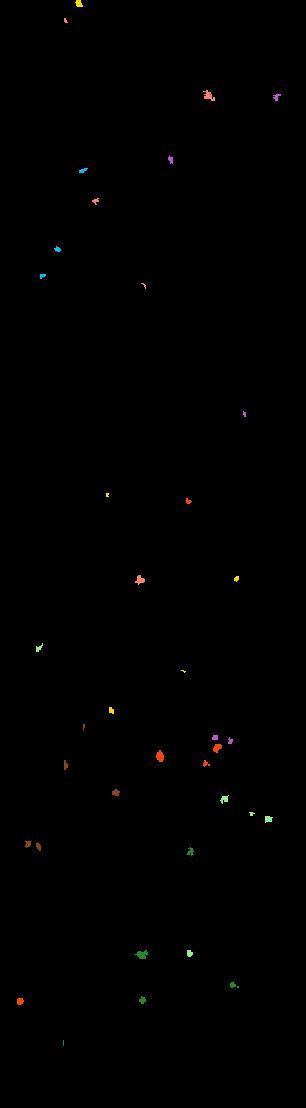}
        \caption{Training Labels}
    \end{subfigure}
    \hspace{0.3em}
    \begin{subfigure}[b]{0.15\textwidth}
        \includegraphics[height=0.46\textheight]{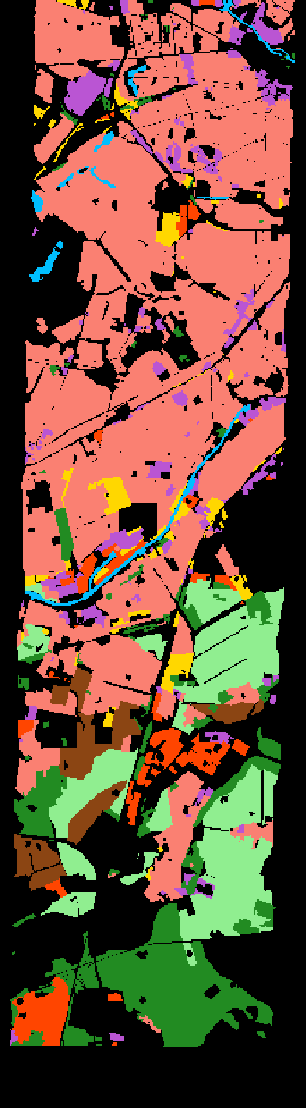}
        \caption{Ground Truth}
    \end{subfigure}
    \hspace{0.3em}
    \begin{subfigure}[b]{0.15\textwidth}
        \includegraphics[height=0.46\textheight]{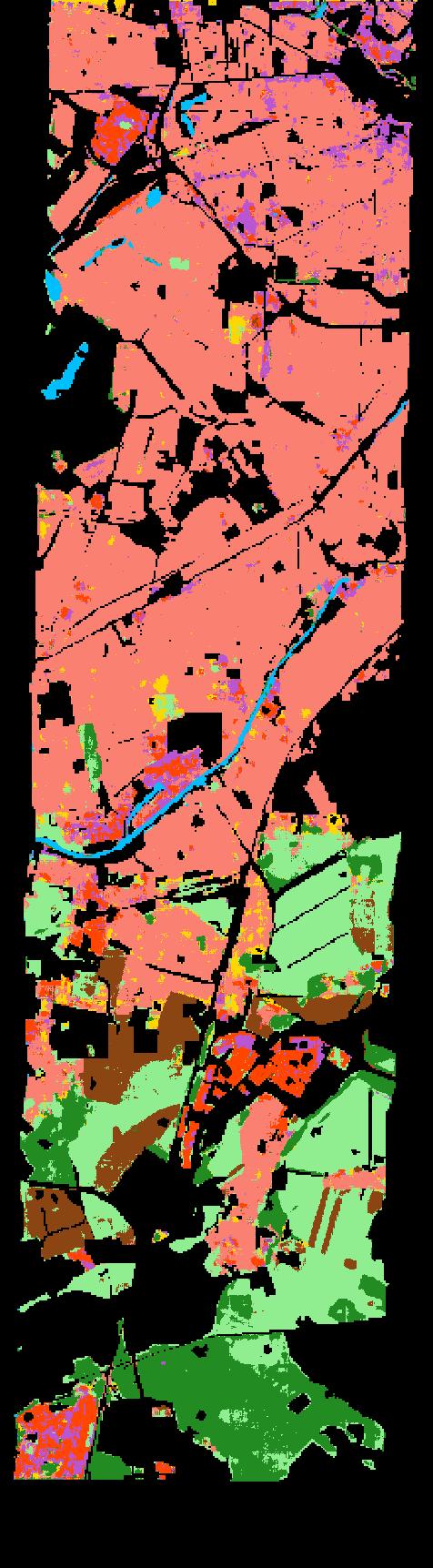}
        \caption{\textbf{UniDiff}}
    \end{subfigure}

    \vspace{0.5em}

    \includegraphics[width=\textwidth]{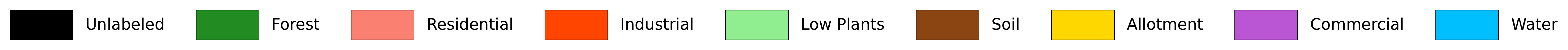}

\caption{Visualization on the Berlin dataset with HSI and SAR. 
(a) Pseudo-RGB HSI, 
(b) PCA-reduced HSI, 
(c) SAR image, 
(d) Sparse Training Labels (limited pixel annotations provided in the benchmark), 
(e) Ground Truth, 
(f) UniDiff (ours, multimodal adaptation with HSI + SAR).}

    \label{Berlin_UniDiff_MM_visual_compara}
\end{figure*}

\begin{figure*}[t]
    \centering
    \begin{minipage}{\textwidth}
        \centering
        \begin{subfigure}[b]{0.161\textwidth}
            \rotatebox{90}{\includegraphics[height=1.01\textwidth]{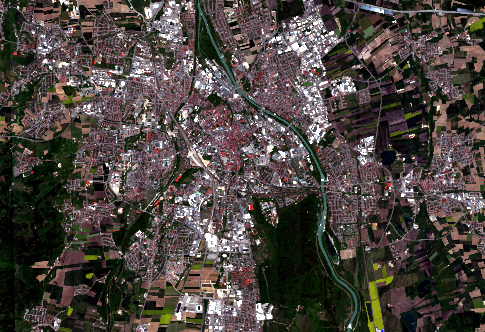}}
            \caption{RGB}
        \end{subfigure}
        \hfill
        \begin{subfigure}[b]{0.161\textwidth}
            \rotatebox{90}{\includegraphics[height=1.01\textwidth]{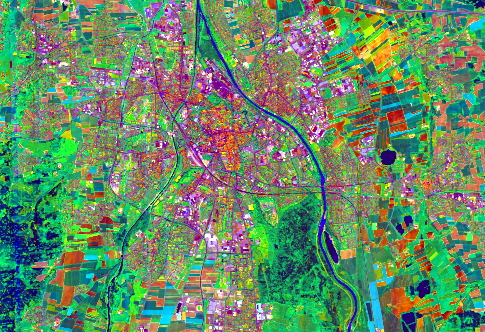}}
            \caption{PCA}
        \end{subfigure}
        \hfill
        \begin{subfigure}[b]{0.161\textwidth}
            \rotatebox{90}{\includegraphics[height=1.01\textwidth]{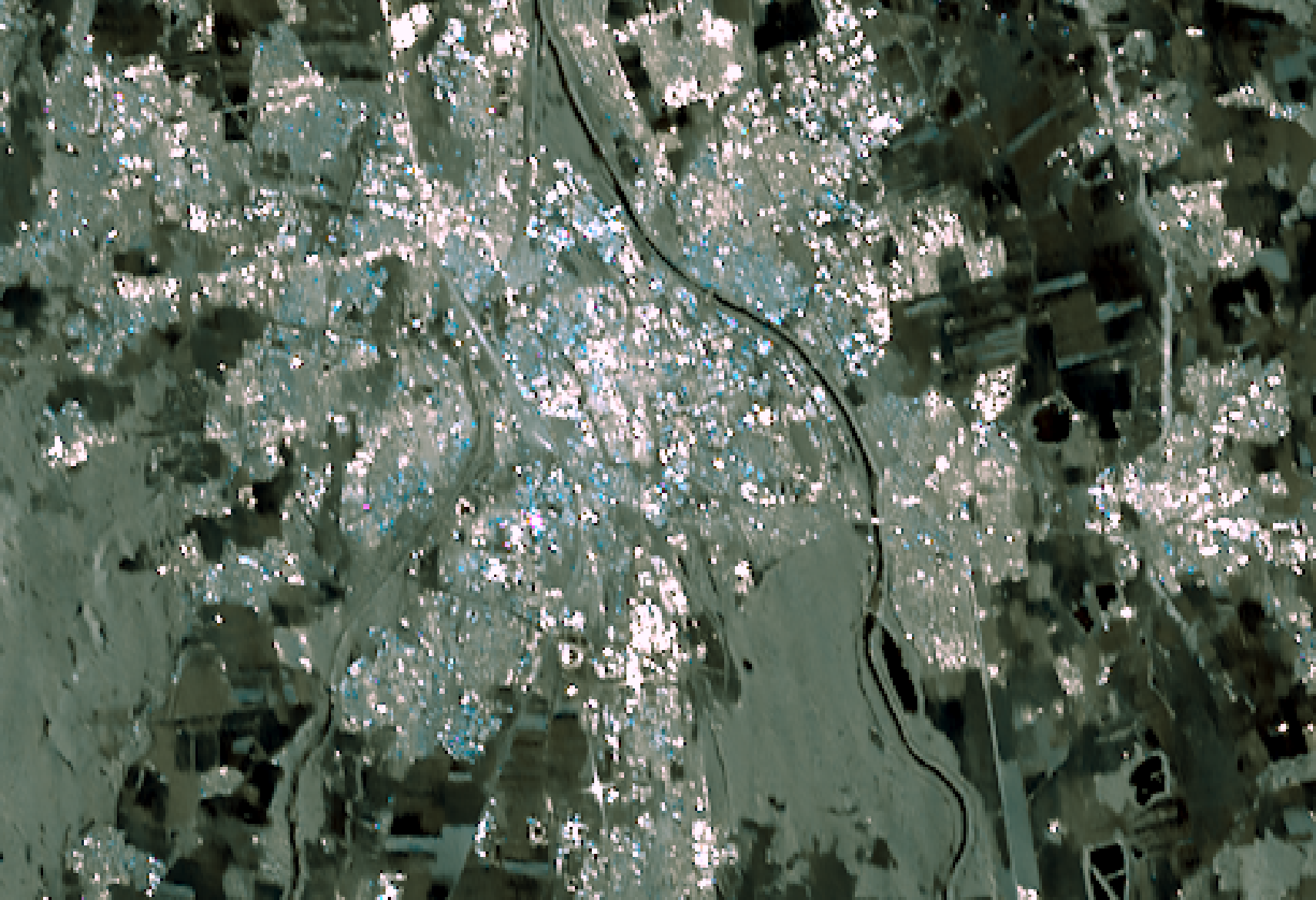}}
            \caption{SAR}
        \end{subfigure}
        \hfill
        \begin{subfigure}[b]{0.161\textwidth}
            \rotatebox{90}{\includegraphics[height=1.01\textwidth]{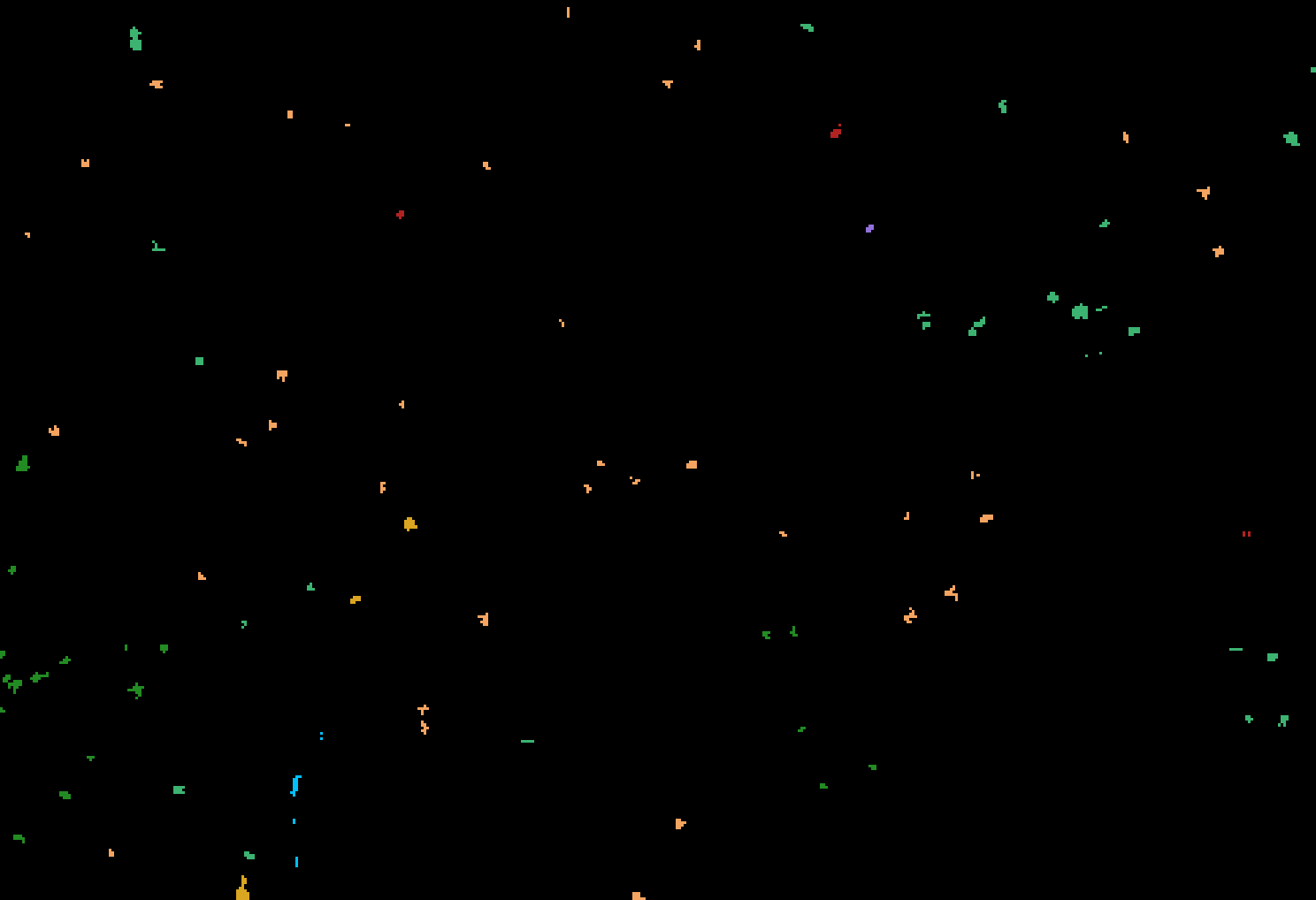}}
            \caption{Training Labels}
        \end{subfigure}
        \hfill
        \begin{subfigure}[b]{0.161\textwidth}
            \rotatebox{90}{\includegraphics[height=1.01\textwidth]{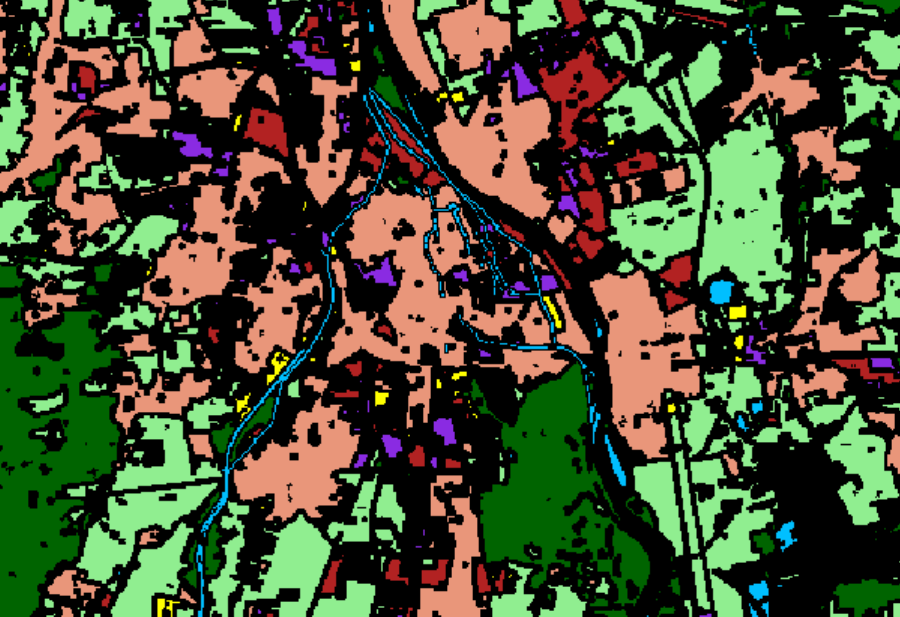}}
            \caption{Ground Truth}
        \end{subfigure}
        \hfill
        \begin{subfigure}[b]{0.161\textwidth}
            \rotatebox{90}{\includegraphics[height=1.01\textwidth]{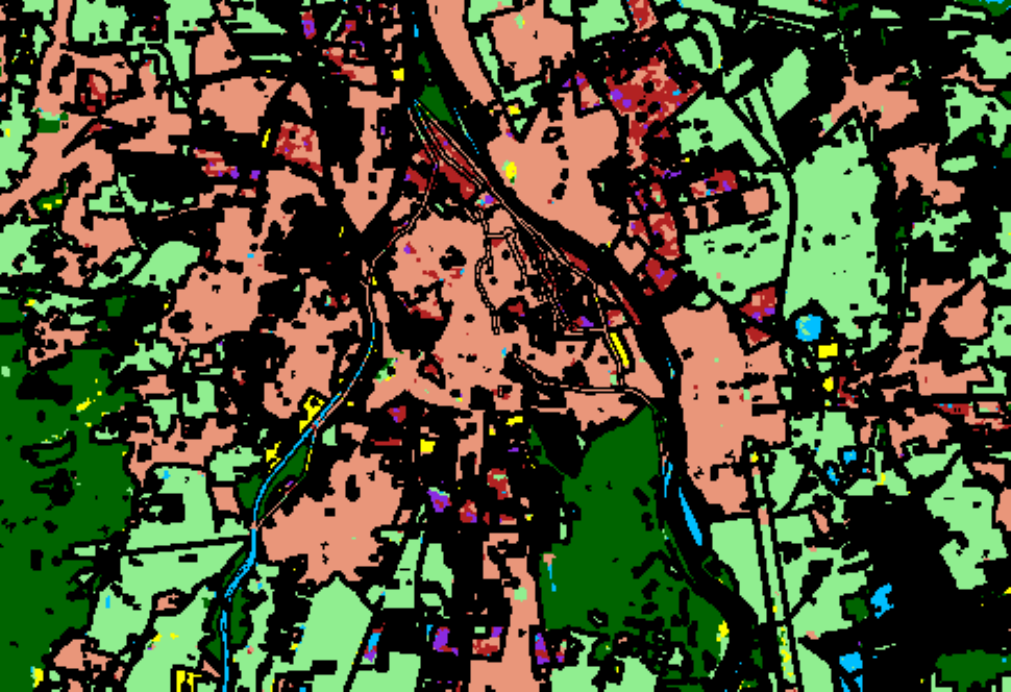}}
            \caption{UniDiff}
        \end{subfigure}
    \end{minipage}

    \vspace{0.5em}
    \includegraphics[width=\textwidth]{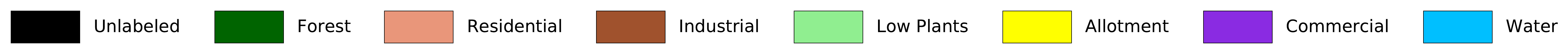}

  % Augsburg (concise)  
\caption{Visualization on the Augsburg dataset. Layout follows \cref{Berlin_UniDiff_MM_visual_compara} (a-c) Input modalities, (d) sparse training labels, (e) ground truth, (f) UniDiff results.}
     \label{UniDiff_Augsburg_Visual}
\end{figure*}

%\vspace{-10pt}
\textbf{Qualitative Results.} 
\cref{Berlin_UniDiff_MM_visual_compara,UniDiff_Augsburg_Visual} present visual comparisons of classification maps. On Berlin, UniDiff reduces salt-and-pepper artifacts in urban areas; on Augsburg, water boundaries are more sharply delineated. Across both datasets, UniDiff preserves spatial detail while improving classification consistency.

\subsection{Ablation Study}
\label{subsec:ablation}
We conduct ablation experiments to assess the effectiveness of multimodal fusion, anchoring strategy, and parameter-efficient adaptation.

\paragraph{Component Analysis (Berlin).}
\cref{tab:ablation_study_berlin} evaluates different modality combinations within our joint adaptation framework. 
Among individual representations, PCA achieves the best result (74.52\% OA), followed by pRGB (73.31\%) and SAR (64.58\%). 
That PCA surpasses pRGB despite lacking textures supports our claim that diffusion backbones leverage spatial layout rather than texture cues (see \cref{sec:intro}). 
Pairwise combinations show mixed outcomes: adding SAR (pRGB+SAR, PCA+SAR at 75.03\%) provides modest gains, while pRGB+PCA (72.97\%) falls below PCA alone. 
The full pRGB+PCA+SAR configuration achieves the strongest performance (80.96\% OA, 70.05\% KC, 66.25\% mF1, 53.34\% mIoU), confirming the benefit of complementary integration across modalities. 
Comprehensive per-class metrics and corresponding Augsburg results are provided in Appendix~C.
\begin{table}[t]
\centering
\caption{Component ablation on the Berlin dataset. 
Unsupervised joint adaptation with all modalities (pRGB, PCA, SAR), 
while evaluation is performed with different input combinations at test time. 
All metrics are reported in percentage. 
pRGB denotes pseudo-RGB from HSI bands.}
\label{tab:ablation_study_berlin}
\small
\setlength{\tabcolsep}{7pt}
\renewcommand{\arraystretch}{1.1}
\begin{tabular}{lcccc}
\toprule
Config & OA & KC & mF1 & mIoU \\
\midrule
pRGB            & 73.31 & 60.05 & 55.60 & 42.33 \\
PCA             & 74.52 & 61.77 & 60.79 & 46.62 \\
SAR             & 64.58 & 47.94 & 48.22 & 34.22 \\
pRGB+PCA        & 72.97 & 60.74 & 61.65 & 47.93 \\
pRGB+SAR        & 75.03 & 63.09 & 61.85 & 48.63 \\
PCA+SAR         & 75.03 & 62.63 & 62.59 & 48.31 \\
\textbf{pRGB+PCA+SAR} & \textbf{80.96} & \textbf{70.05} & \textbf{66.25} & \textbf{53.34} \\
\bottomrule
\end{tabular}
\end{table}

\vspace{-12pt}
\paragraph{Pseudo-RGB Anchoring Strategy.}
We evaluate the role of pseudo-RGB as both a domain anchor and a complementary information source. 
To isolate the anchoring effect, all methods are evaluated using only PCA features for classification. 
As shown in \cref{tab:anchoring_ablation}, PCA-only adaptation leads to catastrophic forgetting, 
performing worse than the pretrained baseline (69.59\% vs 71.27\% OA). 
This suggests that naive adaptation disrupts beneficial ImageNet priors, consistent with catastrophic forgetting~\cite{kirkpatrick2017overcoming}. 
In contrast, joint pRGB+PCA adaptation prevents this degradation and achieves the best results 
(74.96\% OA, +5.37\% over PCA-only). 
These results demonstrate that pseudo-RGB not only provides a complementary spatial HSI representation but also serves as an essential stabilizer during cross-modal adaptation, enabling more reliable pixel-wise classification.

\begin{table}[!h]
\centering
\caption{Pseudo-RGB anchoring ablation on Berlin (T300). 
All methods are evaluated with PCA features only. 
PCA-only trains on PCA; pRGB+PCA trains jointly but is tested on PCA. 
}
\label{tab:anchoring_ablation}
\setlength{\tabcolsep}{6pt}
\renewcommand{\arraystretch}{1.05}
\resizebox{\columnwidth}{!}{%
\begin{tabular}{lccc}
\toprule
Method & OA (\%) & mF1 (\%) & mIoU (\%) \\
\midrule
Pretrained (no adaptation) & 71.27 & 58.89 & 43.88 \\
PCA-only adaptation        & 69.59 & 57.20 & 42.88 \\
pRGB+PCA joint adaptation  & \textbf{74.96} & \textbf{61.44} & \textbf{47.78} \\
\midrule
\textit{$\Delta$ vs PCA-only} & +5.37 & +4.24 & +4.90 \\
\bottomrule
\end{tabular}}
\vspace{-0.3cm}
\end{table}

\begin{figure}[t]
\centering
\setlength{\tabcolsep}{1pt}
\renewcommand{\arraystretch}{0.0}
\setlength{\fboxsep}{0pt}
\setlength{\fboxrule}{0.2pt}
\begin{tabular}{cccccc}
\fbox{\includegraphics[width=0.15\linewidth,height=0.15\linewidth]{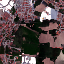}} &
\fbox{\includegraphics[width=0.15\linewidth,height=0.15\linewidth]{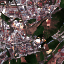}} &
\fbox{\includegraphics[width=0.15\linewidth,height=0.15\linewidth]{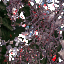}} &
\fbox{\includegraphics[width=0.15\linewidth,height=0.15\linewidth]{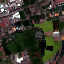}} &
\fbox{\includegraphics[width=0.15\linewidth,height=0.15\linewidth]{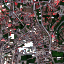}} &
\fbox{\includegraphics[width=0.15\linewidth,height=0.15\linewidth]{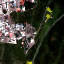}} \\[2pt]
\fbox{\includegraphics[width=0.15\linewidth,height=0.15\linewidth]{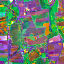}} &
\fbox{\includegraphics[width=0.15\linewidth,height=0.15\linewidth]{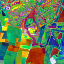}} &
\fbox{\includegraphics[width=0.15\linewidth,height=0.15\linewidth]{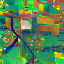}} &
\fbox{\includegraphics[width=0.15\linewidth,height=0.15\linewidth]{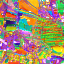}} &
\fbox{\includegraphics[width=0.15\linewidth,height=0.15\linewidth]{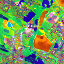}} &
\fbox{\includegraphics[width=0.15\linewidth,height=0.15\linewidth]{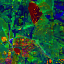}} \\[2pt]
\fbox{\includegraphics[width=0.15\linewidth,height=0.15\linewidth]{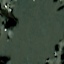}} &
\fbox{\includegraphics[width=0.15\linewidth,height=0.15\linewidth]{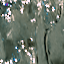}} &
\fbox{\includegraphics[width=0.15\linewidth,height=0.15\linewidth]{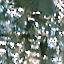}} &
\fbox{\includegraphics[width=0.15\linewidth,height=0.15\linewidth]{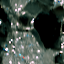}} &
\fbox{\includegraphics[width=0.15\linewidth,height=0.15\linewidth]{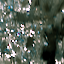}} &
\fbox{\includegraphics[width=0.15\linewidth,height=0.15\linewidth]{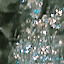}} \\
\end{tabular}
\caption{Representative patches generated after joint adaptation with only 5\% trainable parameters. Rows show pseudo-RGB (top), PCA (middle), and SAR (bottom). The adapted model preserves modality-specific generative characteristics, demonstrating successful parameter-efficient domain adaptation.}

\label{fig:modality_generation}
\vspace{-0.3cm}
\end{figure}

\vspace{-8pt}
\paragraph{Parameter Efficiency Analysis.}
Our approach trains only lightweight adaptive layer norm MLPs ($\sim$5\% of total parameters) while keeping the diffusion backbone frozen. Despite this minimal parameter update, we achieve strong quantitative performance (80.96\% OA on Berlin), demonstrating that pretrained ImageNet knowledge is effectively preserved and leveraged rather than forgotten. \cref{fig:modality_generation} provides visual confirmation: the adapted model generates distinct, realistic patches for each modality—pseudo-RGB maintains natural image characteristics, PCA shows rich spectral patterns, and SAR exhibits appropriate structural features. Together, these quantitative and qualitative results validate that selective parameter updating enables effective domain adaptation without catastrophic forgetting of valuable pretrained priors.

\subsection{Discussion}
\label{subsec:discussion}
\paragraph{Cross-Modal Feature Organization.}
Joint adaptation induces systematic cross-modal organization with dataset-dependent patterns reflecting varying multimodal utility. Performance analysis highlights substantial differences in SAR contribution: Berlin gains +6.00\% OA from HSI+SAR fusion, while Augsburg improves by only +1.08\% (Appendix~B). Cosine similarity analyses (Appendix~D) reveal related organizational patterns—Berlin shows clearer pRGB--SAR separation, whereas Augsburg exhibits more complex reorganization. The most consistent effect across both datasets is reduced within-class variance in cross-modal similarities, indicating that joint adaptation yields more reliable feature relationships, independent of individual modality contribution levels.
\vspace{-8pt}
\paragraph{Limitations.} 
Our approach relies on empirically-determined adaptation stopping criteria and manual selection of diffusion parameters (timesteps and U-Net layers), which are validated across our datasets but may require re-tuning for different sensor modalities or tasks, reflecting broader methodological challenges in cross-modal adaptation and diffusion-based feature extraction. Developing adaptive strategies for stopping and parameter selection remains an exciting direction for future research.
An anonymized implementation is provided at: \url{https://github.com/hutuhehe/UniDiff-code}.

{
    \small
    \bibliographystyle{ieeenat_fullname}
    \bibliography{thesis_bib}
}

\clearpage

\section*{Acknowledgments}
%This work was supported by [funding agency, grant number]. 
We thank Ziyue Xu for error analysis, verification of geographic correspondence using mapping tools, and refinement of visual illustrations, which improved the clarity of the results presented in this work.
We are especially grateful to the anonymous reviewer whose thoughtful guidance on broadening the scope and strengthening the validation helped clarify promising future directions for this work. We also thank the other reviewers for their insightful and constructive feedback.

\end{document}